\documentclass{article}
\usepackage[utf8]{inputenc}
\usepackage{amssymb,amsfonts,amsmath}
\usepackage{graphicx}
\usepackage{fullpage}
\usepackage[super]{natbib}
\usepackage[affil-it]{authblk}

\title{Space as an invention of biological organisms}
\author{Alexander V. Terekhov}
\affil{Institut des Systèmes Intelligents et de Robotique, UMR 7222, CNRS and Université Pierre et Marie Curie, Paris, France}
\author{J. Kevin O'Regan}
\affil{Laboratoire Psychologie de la Perception, UMR 8158, CNRS and Université Paris Descartes, Paris, France}
\date{8 August 2013}

\begin{document}
\maketitle

\textbf{The question of the nature of space around us has occupied thinkers since the dawn of humanity, with scientists and philosophers today implicitly assuming that space is something that exists objectively. Here we show that this does not have to be the case: the notion of space could emerge when biological organisms seek an economic representation of their sensorimotor flow. The emergence of spatial notions does not necessitate the existence of real physical space, but only requires the presence of sensorimotor invariants called `compensable' sensory changes \cite{Poincare1905,Philipona2003,philipona2003perception}. We show mathematically and then in simulations that naïve agents making no assumptions about the existence of space are able to learn these invariants and to build the abstract notion that physicists call rigid displacement, which is independent of what is being displaced. Rigid displacements may underly perception of space as an unchanging medium within which objects are described by their relative positions. Our findings suggest that the question of the nature of space, currently exclusive to philosophy and physics, should also be addressed from the standpoint of neuroscience and artificial intelligence.} 

To illustrate the principle, consider first the sensory universe or ``Merkwelt'' (cf von Uexk\"ull \cite{Uexkull1957}) of the one-dimensional agent in Fig 1. Assume (though this is not known to the agent's brain) that its body is composed of a single photoreceptive sensor that can move laterally inside its body using a ``muscle'' (Fig 1A). Assume a one-dimensional environment as in Fig 1B, and assume first that it is static. If the agent were to perform scanning actions with the muscle and were to plot photoreceptor output against the photoreceptor's actual physical position, it would obtain a plot such as Fig 1D. But it cannot do this because it has no notion, let alone any measure, of physical position, and only has knowledge of proprioception. The agent can only plot photoreceptor output against proprioception, and so obtains a distorted plot as in Fig. 1F. This ``sensorimotor contingency''\cite{mackay1962theoretical,o2001sensorimotor} is all that the agent knows about. It does not know anything about the structure of its body and sensor, let alone that there is such a thing as space in which it is immersed. Indeed the agent does not need such notions to understand its world, since its world is completely accounted for by its knowledge of the sensorimotor contingency it has established by scanning.

But now suppose that the environment can move relative to the agent, for example taking Fig 1B to Fig 1B'. The previously plotted sensorimotor contingency will no longer apply, and a different plot will be obtained (e.g. Fig 1F'). The agent goes from being able to completely predict the effects of its scanning actions on its sensory input, to no longer being able to do so.

However, there is a notable fact which applies. Although the agent does not know this, physicists looking from outside the agent would note that if the displacement relative to the environment is not too large, there will be some overlap between the physical locations scanned before and after the displacement. In this overlapping region, the sensor occupies the same positions relative to the environment as it occupied before the displacement occurred. Since sensory input depends only on the position of the photoreceptor  relative  to the environment, the agent will thus discover that for these positions the sensory input from the photoreceptor will be the same as before the displacement.

Registering such a coincidence is not uncommon for an agent with a single photreceptor, but the same would happen for an agent with numerous receptors. For such a more complicated agent the coincidence would be extremely noteworthy.

In an attempt to better ``understand'' its environment, the agent will thus naturally make a catalogue of these coincidences (cf. arrows in Fig 1F, F'), and so establish a function \(\varphi\) linking the values of proprioception observed before a change to the corresponding values of proprioception after the change. Such a function for all values of proprioception is shown in Fig 1H.

Assume that over time, the environment displaces rigidly to various extents, with the agent located initially at various positions. Furthermore, assume that such displacements can happen for entirely different environments (e.g. Fig 1C). Since the sensorimotor contingencies themselves depend on all these factors, it might be expected that different functions \(\varphi\) would have to be catalogued for all these different cases. Yet it is a remarkable fact that the set of functions \(\varphi\) is much simpler: for a given displacement of the environment, the agent will discover the \emph{same} functions \(\varphi\), even when this displacement starts from different initial positions, and even when the environment is different. 

We shall see below that this remarkable simplicity of the functions \(\varphi\) \emph{provides the agent with the notion of space}. But first let us see where the simplicity derives from.

Each function \(\varphi\) links proprioceptive values before an environmental change to proprioceptive values after the change, in such a way that for the linked values the outputs of the photoreceptor match before and after the change. Seen from outside the agent, the physicist would know that this situation will occur if the agent's photoreceptor occupies the same position relative to the environment before and after the environmental change. And this will happen \emph{if (1) the environment makes a rigid displacement, and if (2) the agent's photoreceptor makes a rigid displacement equal to the rigid displacement of the environment}. Thus physicists looking at the agent would know that the functions \(\varphi\) actually measure, in proprioceptive coordinates, rigid physical displacements of the environment relative to the agent (or  vice versa). (see Supp Mat)

Now we can understand why the set of functions phi is so simple: it is because a defining property of rigid displacements is that they are independent of their starting points, and independent of the properties of what is being displaced.

The functions \(\varphi\) can thus be seen as perceptual constructs equivalent to physical rigid displacements, or one could say, following Jean Nicod\cite{Nicod1923}, that they are \emph{sensible rigid displacements}, where \emph{sensible} refers to the fact that they are \emph{defined within the Merkwelt of the agent}. To illustrate that these sensible rigid displacements or functions phi really have the properties of real physical rigid displacements we will use computer simulations with the more complex two-dimensional agent described in Figure 2. In the Supplementary Material we show that the demonstration applies to an arbitrarily complicated agent, with certain restrictions.

In Figure 3 the two-dimensional simulated agent is first shown an environment that makes a certain displacement (or the agent makes an equivalent hop relative to the environment). The agent is then shown two other instances of the same displacement, but with two completely different environments. Even though in each case the sensory experiences of the agent are different, and even though they change in different ways, the Figure shows that \emph{the same function} \(\varphi\) can be used to account for these changes. This is what is to be expected from a notion of rigid displacement, which should not depend on the content of what is displaced. 

Figure 4 shows further that, once equipped with the notion of sensible rigid displacement, the agent is well on its way towards understanding space. In particular, sensible rigid displacements endow the agent with the percept of space as an \emph{unchanging medium}, which implies being able to distinguish between the sensory changes caused by the proper movements of the agent from those reflecting the deformation of the environment. Figure 4 shows how the simulated agent is able to distinguish between the two despite the fact that in the sensory input a rigid shift may look like a deformation (Figure 4B), while deformations may seem just like a minor displacement (Figure 4C).

Figure 5 shows that the agent can define the notion of \emph{relative position} of A with respect to B. This notion is more abstract than displacement, as there exist numerous paths leading from B to A, while relative position is independent of the choice of a particular path. The notion of relative position allows the agent to ``understand'' that it is at the same ``somewhere'' independently of how it got there. To define the notion of relative position the agent must be able to take different combinations of displacements having the same origin and destination, and consider them as equivalent.

We have shown that, without assuming a priori the existence of space, the agent invents the notions of \emph{sensible displacement}, \emph{unchanging medium} and \emph{relative position}. These notions allow the agent to conceive of its environment in a way that we can assimilate to possessing the notion of space. The agent can now separate the properties of its sensed environment into properties a physicist would call \emph{spatial} (position, orientation) and \emph{non-spatial} (shape, colour, etc). These are the properties whose changes the agent can and cannot account for in terms of \emph{sensible rigid displacements}. Several further points should be mentioned.

The method that the agent uses to ``invent'' its notion of space involves defining \(\varphi\) functions from matching sensory signals. As is the case for temporal coincidence, this can be understood as a strategy of associating causes that lead to the same consequences\cite{markram2011history}. This is a productive learning strategy in general and is easily implementable in neural hardware. 

Note however that constructing sensible rigid displacements on the basis of matches is only possible if sensory changes caused by modifications in the environment can be compensated (i.e. equalized or cancelled) by the agent's own action. The conditions for this to be possible are (1) that the agent be able to act, and (2) that appropriate compensatory changes can occur in the environment. The agent's own actions are thus crucial for the acquisition of the notion of space. Of course if the agent knows in advance that there is space it may be able to reconstruct it without acting. But if the agent has limited action capacities, it will not invent space `correctly'. In particular, the simple two-dimensional agent we have considered has a retina that can translate, but cannot rotate. This agent will therefore classify relative position, but not orientation, as being a spatial property. Evidence from biology also shows the importance of action in the acquisition of spatial notions: an example is the classic result of Held and Hein \cite{Held1963} 

In addition to action, sufficient richness of the environment is essential for an agent to discover space. If for example displacement in a certain direction has no sensory consequences, or if they are ambiguous, then the agent will be unable to learn the corresponding sensible rigid displacements. Again this is coherent with biology, where it has been shown\cite{Blakemore1970} that kittens raised in visual environments composed of vertical stripes are blind to displacements of horizontally aligned objects and vice versa.

Another point worth mentioning is the fact that sensible rigid displacements are nothing but abstract constructs -- they do not imply that something really moves: if the agent inhabited a different physical universe but where the sensorimotor regularities were the same, then it would develop the same construct of sensible rigid displacement. For example in the Supplementary Material we describe an agent whose world consists only of sounds, but that develops sensible rigid displacements in pitch analogous to the spatial constructs of the agent in Figure 1.

A final point concerns the statistical approaches often used up to now to understand brain functioning\cite{ganguli2012compressed,zhaoping2006theoretical}. Such approaches use statistical correlations to compress the data observed in sensory and motor activity. It is possible that these approaches may be adapted to capture the ``algebraic'' notion of mutual compensability between environment changes and an agent's actions that is instantiated by the functions \(\varphi\) and that is essential for understanding the essence of space.

In conclusion, the three-dimensional space we perceive could be nothing but a construct which simplifies the representation of information provided by our limited senses in response to our limited actions. In reality space -- if it exists -- may have a higher number of dimensions, most of which we perceive as non-spatial properties because of our inability to perform corresponding compensatory movements. Or, conversely, there may in fact be no physical space: our impression that space exists may be nothing but a gross oversimplification generated by our perceptual systems, with the real world only being very approximately describable as a collection of ``objects'' moving through an ``unchanging medium''.

\bibliographystyle{plain}
\bibliography{biblio}

\section*{Materials}

\paragraph{Agent.} The two-dimensional agent from Figure 2 was simulated to illustrate the acquisition of spatial knowledge. The agent has a square body in a form of a tray, within which a square retina translates. We choose the measurement units so that the retina movements are confined to a unit square. The position \(x\), \(y\) of the retina is registered by proprioceptors scattered over the body surface and having outputs
\[
p_j = \exp\left\{-\frac{(d_j^p)^2}{(\sigma_j^p)^2}\right\},
\]
where \(d_j^p\) is the distance between the center of the retina and the location of the $j$-th proprioceptor, and \(\sigma_j^p\) is its acuity.

The retina is covered with photoreceptors, measuring the intensity of the light coming from \(N_\ell\) spot light sources located in a plane above the agent. The response of $j$-th photoreceptor is
\[
s_j = \sum_{i=1}^{N_\ell}I_i \exp\left\{-\frac{(d_{ij}^s)^2}{(\sigma_j^s)^2}\right\},
\]
where \(d_{ij}^s\) is the distance between the projection of the \(i\)-th light source onto the plane of the agent and the \(j\)-th photoreceptor; \(I_i\) is the intensity of the \(i\)-th light source, and \(\sigma_j^s\) is the acuity of the \(j\)-th photoreceptor.

For the simulations presented in the paper we deliberately distributed the eight proprioceptors over the agent's body in a non-uniform way so as to ensure a certain amount of distortion of the image in Figure 3-5. Their acuities \(\sigma_j^p\) was set to 0.3 for all receptors. The positions of the nine photoreceptors were drawn randomly from a square with sides of length 0.3. The acuity of the receptors \(\sigma_j^s\) took random values between 0.03 and 0.3. Due to the retinal mobility the agent's `field of view' was a 1.0\(\times\)1.0 square centered at what we call the agent's position.

\paragraph{Learning $\varphi$'s.} The agent was placed into the environment with 200 light sources distributed randomly in 3\(\times\)3 square, centered at the agent's initial position (see Figure S1). The agent scanned the environment by moving the retina inside the body and tabulating the tuples of proprioceptive and photoreceptive inputs \(\langle p_k, s_k\rangle\).  The agent then jumped to a new position, which was within a 1.8\(\times\)1.8 square centered at its initial position, and again scanned the environment and tabulated the tuples \(\langle p'_k, s'_k\rangle\). The agent then looked for the co-occurrences \(s_k=s'_{k'}\) and put the corresponding proprioceptive inputs into pairs \(\langle p_{k}, p'_{k'}\rangle\). The function \(\varphi\) was then defined as the set of all such pairs.

Exclusively for the sake of code optimization when `scanning' the environment the retina moved over a regular 201\(\times\)201 grid. The outputs of the photoreceptors were considered as matching if for every photoreceptor the difference of the outputs before and after the jump was less than 0.005. The corresponding values of proprioception before and after the jump were taken to form the function \(\varphi\). If the value of every proprioceptor in one pair differed by less than 0.01 from the value of proprioceptor in the other pair, then one of the pairs was discarded. The destination points of the agent's jumps also belonged to a regular grid centered at the agent's initial position and having a step size of 0.02. In total we obtained 8281 different functions \(\varphi\).

\paragraph{Sensible rigid displacement.} The agent was facing 40 light sources distributed uniformly along a circle with 0.1 radius. The center of the circle was chosen randomly within a 1.0\(\times\)1.0 square centered at the agent. In the reference displacement all stars moved as a whole to a new random position, which was also within a 1.0\(\times\)1.0 square. The agent determined the function \(\varphi\) corresponding to the reference displacement. Then the agent was shown one of four objects shown in Figure 3: the same circle, a square (composed of 40 lights), a triangle (39 lights), or a star (40 lights). The square and triangle had sides of length  0.2, the star had a ray length 0.3. The objects underwent a random test displacement with initial and final positions within a 1.0\(\times\)1.0 square. In order to save simulation time we only considered displacements which differed from the reference by no more than 0.1 for each axis. The agent determined the functions \(\tilde\varphi\) for each of the tests and computed the distance between \(\varphi\) and \(\tilde\varphi\) as
\begin{equation}
\rho(\varphi, \tilde\varphi) = \sum_{k',\tilde k'\colon p_k = \tilde p_{\tilde k}} \|p'_{k'} - \tilde p'_{\tilde k'}\|,
\label{dist}
\end{equation}
where \(\|\cdot\|\) is a euclidean distance in the proprioception space. The agent identified two displacements as the same if the error was below a threshold which was chosen so that 90\% of displacements of size less than 0.005 were considered identical. The procedure was repeated 1,000 times.

\paragraph{Unchanging medium.} The agent was facing 40 light sources distributed uniformly over a circle with radius 0.1. The center of the circle was chosen randomly in a 0.4\(\times\)0.4 square centered at the agent. The agent scanned the environment and tabulated the tuples \(\langle p_k, s_k\rangle\). Then the agent made a jump to a random point located in a 0.6\(\times\)0.6 square and simultaneously the circle was randomly stretched or shrunk by up to 50\% along a fixed axis. Only those jump destinations were considered for which the agent could `see' the entire circle. The agent scanned the environment again and tabulated new tuples \(\langle p'_k, s'_k\rangle\). The agent then searched for a function \(\varphi\) which gave the best fit of the photoreceptors after the jump based on their values before the jump. In particular, the following error was computed:
\[
\varepsilon = \sum_{k=1}^{N_k}\|s_k-s'_{k'}\|,
\]
where \(k'\) was such that \(\varphi(p_k) = p'_{k'}\). If the error was below the threshold the agent assumed that the environment did not change during the jump. The threshold value of the error was chosen in such a way that the agent answered correctly in 90\% of cases when the deformation of the circle was below 0.5\%. Figure 2 shows the result of simulations computed on the basis of 10,000 repetitions of the test.

\paragraph{Relative position.} The agent was facing an environment filled with 200 light sources with random locations and intensities. It was displaced from its original position to the destination point, which had coordinates (0.6, 0.6) relative to the agent's initial position. The agent determined the reference function \(\varphi_{\text{ref}}\) which gave the best account of the displacement-induced changes of the photoreceptor outputs. Then the environment was replaced with a new randomly generated environment and the agent was moved along a path composed of several segments. At every intermediate point along the path, the agent determined the function \(\varphi_j\) accounting for the changes in photoreceptor values. The agent then computed the composition function \(\varphi_{\text{comp}} = \varphi_n\circ\dots\circ\varphi_1\), where \(n\) is the number of path segments. For any two functions \(\varphi\) and \(\tilde\varphi\) defined by sets of pairs \(\langle p_k, p'_{k'}\rangle\) and \(\langle \tilde p_{\tilde k}, \tilde p'_{\tilde k'}\rangle\) the composition \(\tilde\varphi\circ\varphi\) was defined as a set of pairs \(\langle p_k, \tilde p'_{\tilde k'}\rangle\), such that \(p'_{k'} = \tilde p_{\tilde k}\). The distance between \(\varphi_{\text{ref}}\) and \(\varphi_{\text{comp}}\) was computed using formula 1. The test and reference paths were assumed to correspond to the same relative position if the distance was below the same threshold as for the sensible rigid displacements. The procedure was repeated 1,000 times for two-, three-, and four-segment paths. Each intermediate point of the path was within the 0.9\(\times\)0.9 square centered at the original position. In order to reduce simulation time the final points of all paths lay on the same line and were not more than 0.1 away from the origin.

\clearpage
\newpage
\section*{Figures}

\begin{figure}[h!]
  \includegraphics{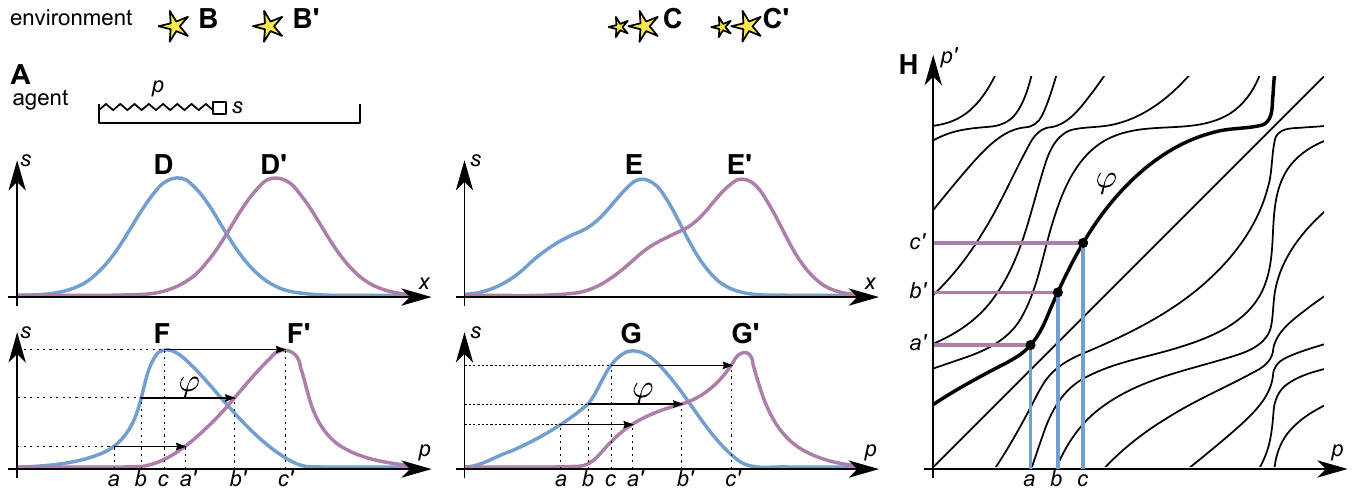}
  \caption{\label{fig1}
Algorithm of space acquisition illustrated with a simplified agent. The agent (A) has the form of a tray, inside which a photoreceptor $s$ moves with the help of a muscle, scanning the environment (B) composed of scattered light sources. The length of the muscle is linked to the output of the proprioceptive cell $p$ in a systematic, but unknown way. The output of the photoreceptor depends on its position $x$ in real space (D). The agent learns the sensorimotor contingency (F) linking $p$ and $s$. After a rigid displacement of the agent, or a corresponding displacement of the environment from B to B', the output of the photoreceptor changes from D to D' and a new sensorimotor contingency F' is established. For a sufficiently small rigid displacement the outputs of the photoreceptor will overlap before and after the displacement. The agent makes a record of the corresponding proprioceptive values between the sensorimotor contingency F and F' (arrows from $a$,$b$,$c$ to $a'$,$b'$,$c'$) and constructs the function $p' = \varphi(p)$ (H, bold line). Different functions $\varphi$ (thin lines in H) correspond to different rigid displacements. If the agent faces a different environment C and makes a rigid displacement equivalent to its displacement to C', the outputs of the photoreceptor change from E to E' and the corresponding sensorimotor contingency changes from G to G'. Yet, because of the existence of space, the same function $\varphi$ links G to G'. The tests in Figure 3-5 show that the functions $\varphi$ constitute the basis of spatial knowledge.
}
\end{figure}

\begin{figure}[ht!]
  \includegraphics{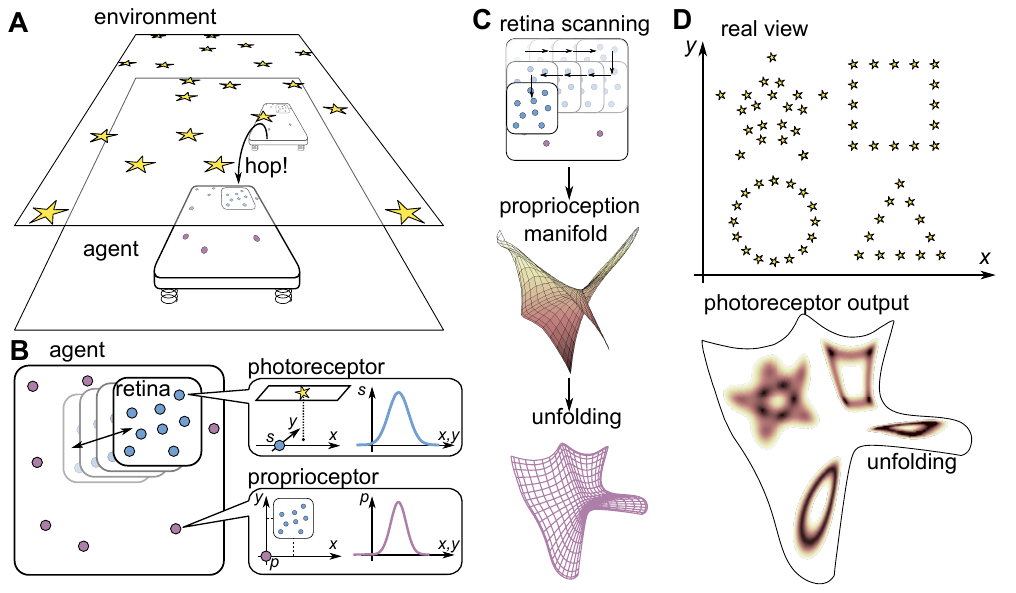}
  \caption{\label{fig2}
The tray-like two-dimensional agent. (A) The agent inhabits a plane where it can perform uncontrolled hops (or equivalently, the environment can shift through unknown distances), resulting in the translation of the agent's body in an unknown direction through an unknown distance. Outside of the agent's plane there is an environment made of light sources. The agent can sense the light sources with nine photoreceptors placed on its mobile retina (B), which can translate with the help of muscles, and whose position is sensed by eight pressure-sensitive proprioceptive cells scattered over the agent's body. As the retina performs the scanning motion (C) the proprioceptors take values lying in a two-dimensional proprioceptive manifold inside the eight-dimensional space of the possible proprioceptive outputs. This manifold can be unfolded into a plane. (D) An example environment and the output of one photoreceptor over this unfolding as the agent performs scanning movements of the retina. This unfolding will be used hereafter in order to illustrate the outputs of the photoreceptors as the agent performs scanning movements of the retina.
}
\end{figure}

\begin{figure}[ht!]
  \includegraphics{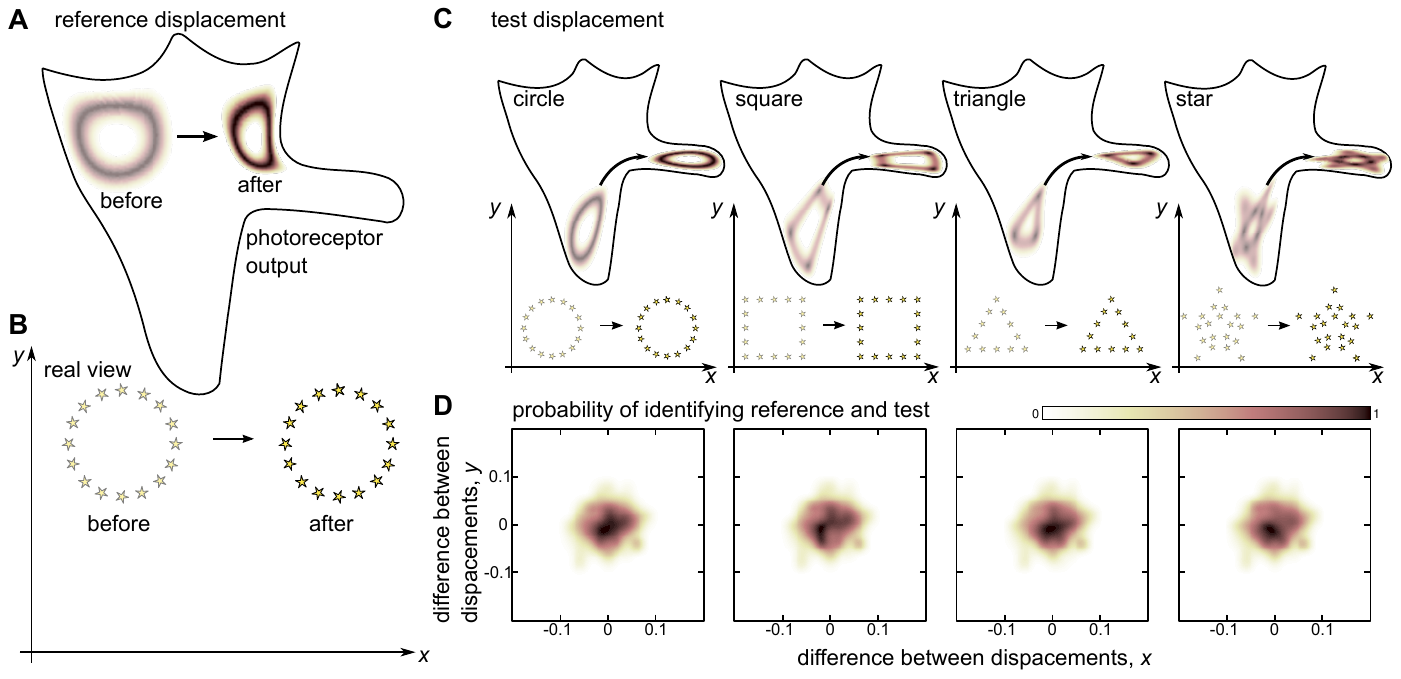}
  \caption{\label{fig3}
The notion of sensible rigid displacements. Seemingly different changes in sensory input will be associated if they correspond to the same displacement in real physical space. The 2d agent is presented with a reference displacement of the environment (B), which it scans before and after the displacement. The output of one of the photoreceptors over the unfolded proprioceptive manifold (Figure 2C) is presented in (A). Then the agent is presented with test displacements (C) from different initial positions and for different environments. Even though the test displacements may strongly alter the shape of the reference, the agent succeeds in associating test and reference if they correspond to the same physical displacement (D). This ability of the agent provides the basis of the notion of \emph{displacement} independent of the environment.
}
\end{figure}

\begin{figure}[ht!]
  \includegraphics{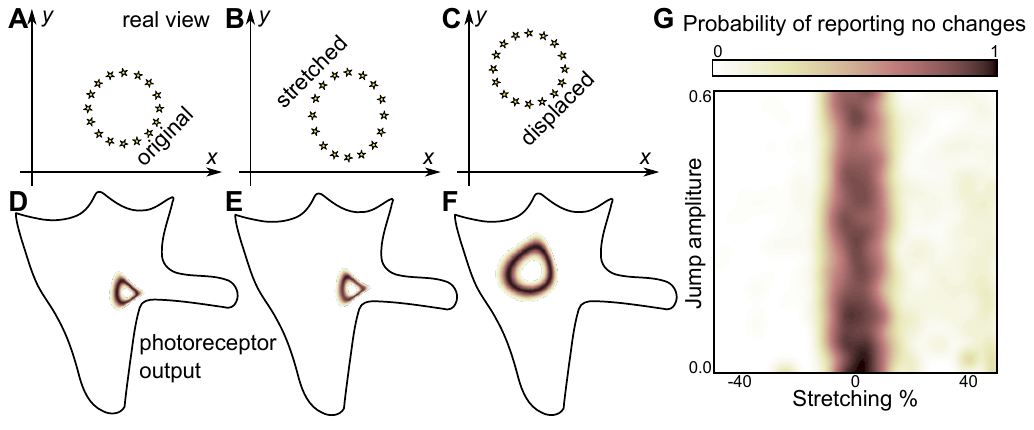}
  \caption{\label{fig4}
The notion of space as unchanging medium. The 2d agent can distinguish between the sensory changes provoked by its own movement and those resulting from the joint effect of its own motion and changes in the environment. The agent is presented with an environment (A) which it scans (D). Then the agent makes a jump and simultaneously the environment is stretched or shrunk along one axis by a certain amount (B), which can be zero (C). The agent is to judge whether the environment was the same before and after the jump. Note that the visual input in the modified environment (E) resembles the original (D) more than it does the unchanged environment (F). Yet the agent can successfully identify the case when the environment does not change, and it can do this independently of the extent of the jumps (G). This ability of the agent underlies the notion of space as an \emph{unchanging medium} through which the agent makes \emph{displacements}.
}
\end{figure}

\begin{figure}[ht!]
  \includegraphics{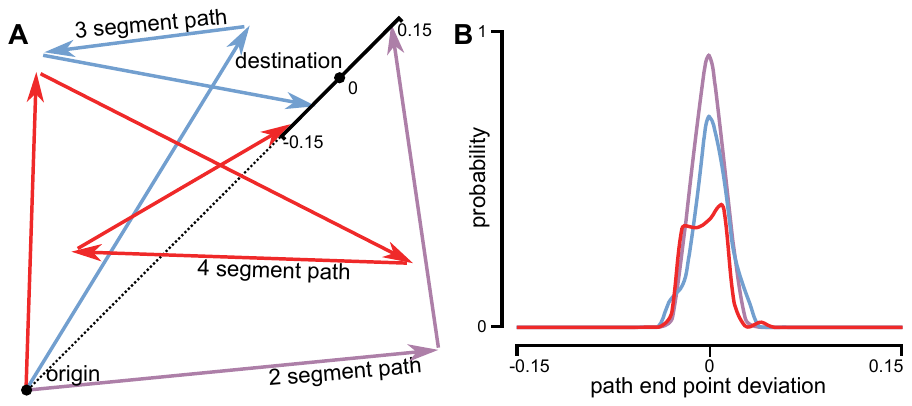}
  \caption{\label{fig5}
The notion of relative position, independent of the connecting path. The 2d agent can construct the notion of \emph{relative position} of a destination point with respect to an origin by associating all possible paths connecting the origin to the destination. The agent was presented with a reference single-segment path. Then it was returned to the origin and moved along two-, three-, or four-segment paths with possibly different destination points at a distance in the range [-0.15, 0.15] from the destination of the reference path; all points were chosen on the same straight line (A). The surrounding environment was altered at every path presentation in order to ensure that the agent calculated the displacement instead of comparing the view at the destination points. The agent successfully associated together paths which arrived close to the original destination point. This association was more accurate for the 2-segment path, and became weaker as the number of path segments increased (B). This can be explained by accumulation of the integration error.
}
\end{figure}

\clearpage
\section*{Supplementary Materials}

\subsection*{Formalization}

Here we consider a general agent immersed in real physical space. Leter we will abandon the assumption of the existence of physcial space and give the conditions for the emergence of perceptual `space-like' constructs independently of whether they correspond to any real physical space.

Let \(s\) be the vector of the agent's exteroceptor outputs. The exteroceptors are connected to a body, assumed to be rigid, whose position and orientation is described by a spatial coordinate defined by the vector \(x\). For every environment \(\mathcal E\) the outputs of the exteroceptors are defined by a function
\begin{equation}
s = \sigma_{\mathcal E}(x).
\label{sigma}
\end{equation}

We assume that this function has the property that if the environment \(\mathcal E\) makes a rigid motion and becomes \(\mathcal E'\), then there exists a rigid transformation \(T\) of entire space such that
\begin{equation}
s' = \sigma_{\mathcal E'}(x) = \sigma_{\mathcal E}(T(x)).
\label{sigma_inv}
\end{equation}

Proprioception \(p\) reports the position of the exteroceptors in the agent's body. For a given position of the agent \(\mathcal X\) we assume there is a function  \(\pi_{\mathcal X}\) such that
\begin{equation}
p = \pi_{\mathcal X}(x).
\label{pi}
\end{equation}

Again, the function \(\pi_{\mathcal X}\) has the property that the agent's displacement to a position \(\mathcal X'\) can be accounted for by the rigid transformation \(\mathcal T\) of entire space:
\begin{equation}
p' = \pi_{\mathcal X'}(x) = \pi_{\mathcal X}(\mathcal T(x)).
\label{pi_inv}
\end{equation}

Assuming that proprioception unambigously defines the position of the exteroceptors in space
\[
x = \pi_{\mathcal X}^{-1}(p)
\]
and
\[
s = \left(\sigma_{\mathcal E}\circ\pi_{\mathcal X}^{-1}\right)(p)
\]
where the function \(\sigma_{\mathcal E}\circ\pi_{\mathcal X}^{-1}\) is the sensorimotor contingency learned by the agent for every position \(\mathcal X\) of itself and of the environment \(\mathcal E\).

When the agent or the environment moves a new sensorimotor contingency is established
\[
s' = \left(\sigma_{\mathcal E'}\circ\pi_{\mathcal X'}^{-1}\right)(p') = \left(\sigma_{\mathcal E}\circ T\circ\mathcal T^{-1}\circ\pi_{\mathcal X}^{-1} \right)(p').
\]

The agent learns the function \(\varphi\) linking the values \(p\) and \(p'\) such that \(s=s'\), or
\[
\left(\sigma_{\mathcal E}\circ\pi_{\mathcal X}^{-1}\right)(p) =  \left(\sigma_{\mathcal E}\circ T\circ\mathcal T^{-1}\circ\pi_{\mathcal X}^{-1} \right)(p').
\]
The function \(\varphi\) is not always defined uniquely since the mapping \(\sigma_{\mathcal E}\) can be non-invertible. It can be inverted in the domain of its arguments if the environment is sufficiently rich, i.e. if the total vector of exteroceptor outputs is different at every position of the exteroceptors within the range admitted by the proprioceptors. In this case
\begin{equation}
\varphi = \pi_{\mathcal X}\circ\mathcal T\circ T^{-1}\circ\pi_{\mathcal X}^{-1}.
\label{phi}
\end{equation}

It can be seen from the expression for the function \(\varphi\) that it simply gives a proprioceptive account of the relative rigid displacement \(\mathcal T\circ T^{-1}\) of the environment and the agent. The functions \(\varphi\) are thus the agent's \emph{sensible rigid displacements}, which are associated with the environment's rigid motion from \(\mathcal E\) to \(\mathcal E'\) and the agent's rigid motion from \(\mathcal X\) to \(\mathcal X'\). As is clear from equation \ref{phi}, the function \(\varphi\) only depends on the transformations \(T\) and \(\mathcal T\). In physical space these transformations depend only on the displacements themselves and are independent of the initial positions of the agent and the environment, and of the content of the environment. Moreover, since the transformations \(T\) and \(\mathcal T\) form Lie groups, the functions \(\varphi\) also inherit some group properies. For any two \(\varphi\) functions
\[
\varphi_1 = \pi_{\mathcal X}\circ\mathcal T_1\circ T_1^{-1}\circ\pi_{\mathcal X}^{-1} \quad\text{and}\quad \varphi_2 = \pi_{\mathcal X}\circ\mathcal T_2\circ T_2^{-1}\circ\pi_{\mathcal X}^{-1}
\]
there exists a function \(\varphi_3\) such that
\[
\varphi_1 \circ \varphi_2 = \pi_{\mathcal X}\circ\mathcal T_1\circ T_1^{-1}\circ\mathcal T_2\circ T_2^{-1}\circ\pi_{\mathcal X}^{-1} = \pi_{\mathcal X}\circ\mathcal T_3\circ T_3^{-1}\circ\pi_{\mathcal X}^{-1} = \varphi_3
\]
where \(\mathcal T_3\) and \(T_3\) are transformations describing the total displacements of agent and of the environment, for which \(\mathcal T_3\circ T_3^{-1} = \mathcal T_1\circ T_1^{-1}\circ\mathcal T_2\circ T_2^{-1}\).

The functions \(\varphi\) do not form a group. This is because they are defined only on a subset of proprioceptive values, for which the exteroceptor outputs overlap before and after the shift. It may happen that the domain of definition of the function \(\varphi_3\) is larger than that of \(\varphi_1\circ\varphi_2\) and hence the composition \(\varphi_1\circ\varphi_2\) is not one of the functions \(\varphi\).

\medskip

Up until this point we have assumed the existence of real physical space. Now we would like to abandon this assumption, and only retain the conditions which allow the construction of the function \(\varphi\).
This gives us a list of requirements for the existence of `space-like' constructs. (1) There must be a variable \(x\) and functions \(\sigma_{\mathcal E}\) and \(\pi_{\mathcal X}\) such that the outputs of the extero- and prioprioceptors can be described by the equations \ref{sigma} and \ref{pi}, and the function \(\pi\) must be invertible. The agent must be ably to `act', i.e. induce changes in the variable \(x\). Moreover, (2) there must exist (and be sufficiently often) changes of the environment \(\mathcal E\to\mathcal E'\) and/or of the agent \(\mathcal X\to\mathcal X'\) such that the equations \ref{sigma_inv} and \ref{pi_inv} hold. The corresponding transformations \(T\) and/or \(\mathcal T\) must be applicable to all environments \(\mathcal E\) and external states of the agent \(\mathcal X\), and they must form a group with respect to the composition operator.

Note that the requirement (2) does not presume that there are no other types of changes of the environment and/or of the agent. The agent will identify only the changes possessing such a property as \emph{sensible rigid displacements} and will obtain the functions \(\varphi\) that correspond to them.

Also note again that here we do not assume the existence of space. We only make certain assumptions regarding the structure of the sensory inputs that the agent can receive.

\medskip

The agent presented in Figure 2 of the main text will only recognize translations as the spatial changes, because it can only translate its retina, and hence for this agent the variable \(x\) only includes the position of the retina in space, not it's orientation.

One can imagine an agent that can stretch its retina in addition to translations and rotations. For such an agent the variable \(x\) will include position, orientation and stretching of the retina. If this agent can stretch its entire body, or if the environment has a tendency for such deformations, then stretching will be classified as a \emph{sensible rigid displacement} similarly to translations and rotations.

One can also imagine an agent whose sensory inputs do not depend on physical spatial properties, but satisfy the requirements described above. Such an agent will develop a false notion of space, where it is not present. The description of such an agent is given below.

\subsection*{Audio agent}

Here we show that an agent can develop incorrect spatial knowledge, i.e. that does not correspond to physical space, if the conditions presented in the previous section are satisfied. The agent, inspired by Jean Nicod, inhabits the world of sounds (Figure S2). Its environment is a continuously lasting sine wave, or a chord (Figure S2A). The agent consists of a hair-cell, which oscillates in response to the acoustic waves (Figure S2B). The amplitude of this oscillation is measured by an exteroceptor \(s\). The response is maximal if the frequency \(f\) of one of the sine waves coincides with the eigenfrequency of the hair-cell.

The agent can `scan' the environment by changing the stiffness at the cell's attachment point and thus its eigenfrequency, which is measured by the proprioceptor \(p\). For the environment B the dependency between the amplitude and the cell's eigenfrequency has the shape illustrated in Figure S2D. We assume that for any other note (like B') the dependency between the amplitude and the cell's eigenfrequency remain the same, but shifted (Figure S1D').

The agent does not know these facts. It only knows the dependency between exteroception \(s\) and proprioception \(p\), which consistutes the sensorimotor contingency (Figure S2D)  corresponding to the environment B. For a new note (A') a new sensorimotor contingency D' is established. Yet, as before, the agent notices that the outputs of the exteroceptor \(s\) coincide for certain values of \(p\). It makes note of these coincidences and defines the functions \(\varphi\) corresponding to all changes of the notes (Figure S2H).

The same procedure applies if the agent faces a chord of two  (C and C') or more notes. Instead of changes in the pitch of a note, we now have transposition of the whole chord. The agent can discover that the same set of functions \(\varphi\) works for notes and for chords.

Although this agent is unable to move in space and although it only perceives continuous sound waves, it can nevertheless build the basic notions of space. However these notions are `incorrect', in the sense that they do not correspond to actual physical space, but to the set of note pitches. The sensible rigid displacements for this agent correspond to transpositions of the chords. The unchanging medium is the musical scale, and the relative position of one chord with respect to an identical but transposed chord is just the interval through which the chord has been transposed. For such an agent a musical piece is somehow similar to what a silent film is for us: it is a sequence of objects (notes), appearing, moving around (changing pitch) and disappearing.

Using the formalism introduced above, we can say that for this agent the spatial variable is frequency, \(f\). For any given environment \(\mathcal E\), which in this case is constituted by simultaneously played notes, the output of the exteroceptor \(s\) depends only on the eigenfrequency of the hair-cell, which can be measured using the same variable \(f\). This means that the function \(\sigma_{\mathcal E}(f)\) exists. The rigid shift of the environment \(\mathcal E\) to \(\mathcal E'\), which is the chord transposition, results just in frequency scaling: \(\sigma_{\mathcal E'}(f)\) = \(\sigma_{\mathcal E}(kf)\). Evidently, these transformations form a group. Proprioception \(p\) signals the stiffness of the hair-cell, which is functionally related to its eigenfrequency, and hence the invertible function \(\pi(f)\) also exists. As our auditory agent is unable to perform anything similar to rigid displacements, the function \(\pi\) does not depend on anything equivalent to the state \(\mathcal X\) of our original simple agent (Figure 1).

The existence of the functions \(\sigma_{\mathcal E}(f)\) and \(\pi(f)\) fulfills the requirement (1) from the previous subsection. We can assume that music being played is just a piano exercise and hence the chords are often followed by their transposed versions. In this case there exist (and are sufficiently often) changes of the environment \(\mathcal E\to \mathcal E'\), which correspond to a simple shift of all played notes by the same musical interval. These shifts evidently form a group, and hence the requirement (2) is also fulfilled. The fulfilment of these two requirements suffices for the existence of sensible rigid displacements and thus for basic spatial knowledge, described above.

\begin{figure}
\includegraphics{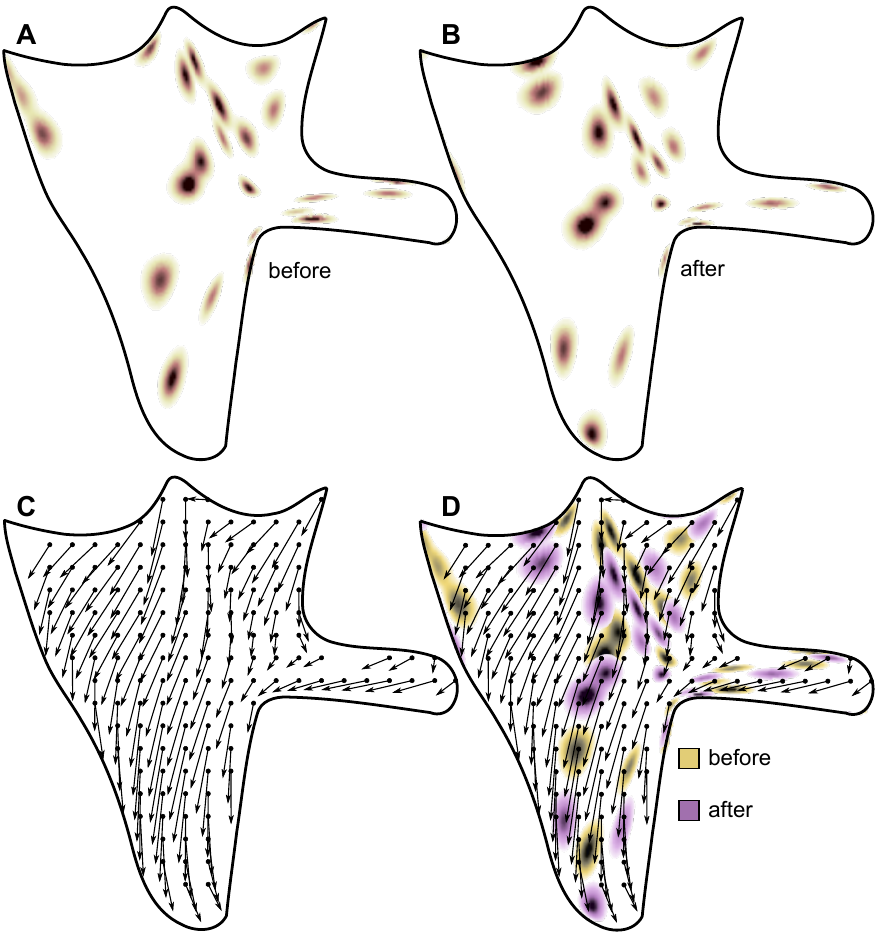}

\textbf{Figure S1:} Acquisition of \(\varphi\) functions. The agent scans the environment composed of random light sources before (A) and after (B) a rigid displacment caused by the agent's jump. The corresponding function \(\varphi\) is illustrated in C with the arrows connecting points of coinciding photoreceptor outputs before (the origin of the arrow) and after the jump (the end of the arrow). D illustrates the meaning of the function \(\varphi\) , which is the field of the proprioceptive changes necessary to compensate changes in photoreceptors induced by the rigid motion (jump).
\end{figure}

\begin{figure}
\includegraphics{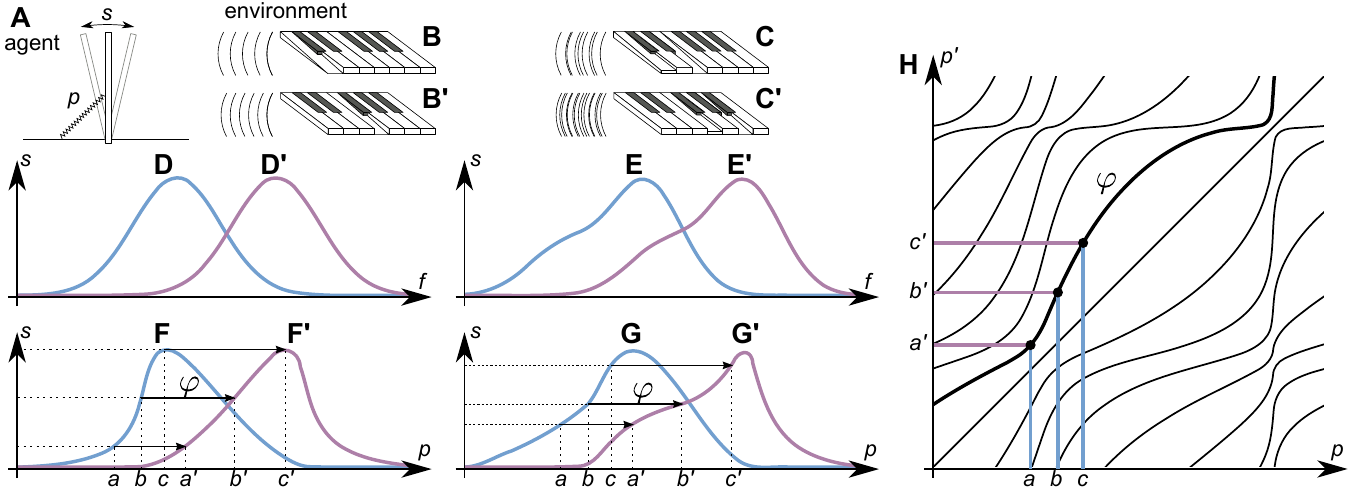}

\textbf{Figure S2:} A simple audio agent. The agent (A) consists of a rod attached to a support, oscillating in response to the acoustic waves (B). The agent measures the intensity of this oscillation with an `exteroceptor' \(s\). By exercising its muscle the agent can change the stiffness of the spring at the support and thus the eigenfrequency \(f\) of the rod, measured with the proprioceptor \(p\). For an acoustic wave generated by a single note (A) the exteroceptor response \(s\) depends on the eigenfrequency \(f\) as shown in D. However the agent measures the proprioceptive response  \(p\) and not the eigenfrequency, and so only has access to the sensorimotor contingency (F). The change of the note from A to A' results in a shift in the characteristics D to D' and in the establishment of a new sensorimotor contingency F'. The agent can learn the functions \(\varphi\) (H) by taking note of the coincidences in the exteroceptor output \(s\). If a pair of notes is played (C and C') with the dependencies E and E', new sensorimotor contingencies are obtained by the agent (G and G'), yet the same functions \(\varphi\) link them together.
\end{figure}

\end{document}